\documentclass{article}

\usepackage{PRIMEarxiv}

\usepackage[utf8]{inputenc} % allow utf-8 input
\usepackage[T1]{fontenc}    % use 8-bit T1 fonts
\usepackage{hyperref}       % hyperlinks
\usepackage{url}            % simple URL typesetting
\usepackage{booktabs}       % professional-quality tables
\usepackage{amsfonts}       % blackboard math symbols
\usepackage{nicefrac}       % compact symbols for 1/2, etc.
\usepackage{microtype}      % microtypography
\usepackage{lipsum}
\usepackage{fancyhdr}       % header
\usepackage{graphicx}       % graphics
\graphicspath{{media/}}     % organize your images and other figures under media/ folder
\usepackage{multirow}

\usepackage{array}

\title{Evaluating Large Language Models on Graphs: Performance Insights and Comparative Analysis
}

\author{
  Chang Liu, Bo Wu \\
  Department of Computer Science \\
  Colorado School of Mines \\
  \texttt{\{liuchang, bwu\}@mines.edu} \\
}

\begin{document}
\maketitle

\begin{abstract}
Large Language Models (LLMs) have garnered considerable interest within both academic and industrial. Yet, the application of LLMs to graph data remains under-explored. In this study, we evaluate the capabilities of four LLMs in addressing several analytical problems with graph data. We employ four distinct evaluation metrics: Comprehension, Correctness, Fidelity, and Rectification. Our results show that: 1) LLMs effectively comprehend graph data in natural language and reason with graph topology. 2) GPT models can generate logical and coherent results, outperforming alternatives in correctness. 3) All examined LLMs face challenges in structural reasoning, with techniques like zero-shot chain-of-thought and few-shot prompting showing diminished efficacy. 4) GPT models often produce erroneous answers in multi-answer tasks, raising concerns in fidelity. 5) GPT models exhibit elevated confidence in their outputs, potentially hindering their rectification capacities. Notably, GPT-4 has demonstrated the capacity to rectify responses from GPT-3.5-turbo and its own previous iterations. The code is available at: \texttt{https://github.com/Ayame1006/LLMtoGraph}.

\end{abstract}

% keywords can be removed
\keywords{Large Language Model \and GPT \and Graph}

\section{Introduction}

Graph, or graph theory, is a fundamental part of numerous areas in the modern world, particularly in technology, science, and logistics \cite{ji2021survey}. Graph data represents the structural characteristics between nodes, thus illustrating the connections within the components of the graph. Many real-world datasets, such as social networks, citation networks, and gene expression networks, are intrinsically represented as graphs.

Large Language Models (LLMs) \cite{ouyang2022training} have seen been rapid development in recent years \cite{zhao2023survey, touvron2023llama}. They have shown high performance in plenty of areas, such as understanding and generating human languages, question answering, and information retrieval \cite{bubeck2023sparks, bang2023multitask, chen2021evaluating, chang2023survey}. In nature language processing, Generative Pre-trained Transformer (GPT) models are among the most prominent instances of LLMs \cite{bubeck2023sparks, openaigpt4, liu2023summary}. Their widespread popularity is substantially due to their remarkable capacity to comprehend the underlying semantics inherent in natural languages.

While LLMs have shown effectiveness across multiple domains \cite{zhao2023survey}, their capability to comprehend and reason with graph data remains insufficiently studied. Evaluating LLMs with graph data can reveal their suitability for graph-based challenges. Such evaluations can extend LLM applicability to complex tasks like relationship detection between entities, multi-hop knowledge inference, and pattern recognition in graphs. Given its role in representing entity relationships, graph data is pivotal for addressing real-world complexities \cite{fernandes2018graph}. Mastering graph data interpretation can significantly bolster the realistic relevance and application of LLMs.

Prior studies have assessed the performance of LLMs on graphs, primarily concentrating on GPT models such as text-davinci-003 \cite{guo2023gpt4graph} and ChatGPT \cite{wang2023can}. The evaluated LLMs predominantly comprise closed-source models developed by OpenAI. Conversely, other state-of-the-art open-source models have been largely overlooked. Furthermore, the majority of these investigations center on the LLMs' ability to comprehend graph-related queries and produce accurate responses. They haven't delved into intricate matters such as fidelity in multi-answer questions or self-rectification capabilities.

In the current investigation, we systematically analyze LLMs' ability to comprehend graph data based on graph topology. Our main goal is to assess LLMs' proficiency in assimilating and processing topological data inherent in graphs, which are transcribed in natural language. We employ multiple prompting techniques to evaluate different LLMs using randomly generated graph data, ensuring unbiased performance assessment. This research provides insights into LLMs' interpretation of graph-based information, contributing to the overarching domains of graph-based AI and natural language processing.

In summary, our main contributions are as follows. 

(1) We formulate a set of queries to evaluate the comprehension of LLMs on graph data. Our main objective is to demonstrate the capacity of LLMs to accurately extract, comprehend, and subsequently analyze the topological constructs inherent within the graph structures. 

(2) This study assesses four LLMs for their proficiency in graph data comprehension and response accuracy. Comparative analysis reveals that two open-source models exhibited substantially lower performance in generating correct answers compared to the two GPT models.

(3) To show models' capabilities in graph topological reasoning, we design a set of specialized tasks targeting this aspect of LLM performance. With increasing demands of graph topological reasoning, the correctness of LLM responses is notably affected.

(4) In the context of graph topological reasoning, advanced techniques like zero-shot chain-of-thought and few-shot prompting do not consistently improve LLM performance. These approaches occasionally lead LLMs to produce spurious outcomes, though GPT-4 appears to be more robust than GPT-3.5-turbo.

(5) In multi-answer tasks, GPT models can produce a considerable quantity of inaccurate responses, questioning their overall fidelity. Notably, GPT-3.5-turbo produces misleading answers at a rate seven times greater than accurate ones in one of the evaluations.

(6) GPT models exhibit elevated confidence in their generated responses, potentially hindering self-rectification. Notably, GPT-4 demonstrates an ability to rectify and calibrate responses from both its predecessor and its own previous iterations.

\section{Problem Definition} 

This study aims to assess the ability of LLMs to tackle problems related to graph structures. While graph data is presented in sequential text, it inherently represents inter-node relationships, signifying a distinct topological structure. As a result, we construct several tasks to perform a thorough comparison of LLM performance. Unless otherwise specified, LLMs are instructed to produce a singular response for multi-answer tasks to mitigate the complexity of enumerating all answers.

(1) \textbf{Connectivity (Find a Simple Path)}: this task involves locating a straightforward path from a source node \textit{p} to a target node \textit{q} within a given graph. The problem is purposefully constructed to assess the capabilities of LLMs in traversing and deciphering the structural information of a graph via its edges. 

(2) \textbf{Node neighbor classification}: the task is delineated to verify whether a node constitutes a \textit{k}-hop neighbor to a target node. This primary problem has been further forked into three sub-questions relating to 1-hop, 2-hop, and 3-hop node neighbor classification, respectively. To obviate potential ambiguity, a node with multiple routes to the target node is classified as a lower-hop neighbor. Such a class of problems is instrumental in comparing the proficiency of LLMs to process data pertaining to one to multiple hop graph structures. 

(3) \textbf{Node degree}: the task determines the degree of a specific target node within a given graph. The goal is to showcase the ability of LLMs to execute comprehensive search operations over a targeted node within the edge list. 

(4) \textbf{Pattern matching}: this task of identifying a wedge or a triangle that encompasses a target node constitutes another significant problem in our study. Specifically, the target node can be defined as the center of the wedge identified by the LLMs. Pattern matching is generally deemed challenging due to its stringent spatial reasoning demands on models. Despite limiting the investigation of patterns to a wedge or a triangle, these questions sufficiently underscore the complexities inherent in structural reasoning tasks. 

(5) \textbf{All shortest paths}: this task is formulated to prompt the LLMs to identify all the shortest paths between a designated source node \textit{p} and a target node \textit{q} within a specified graph. The complexity of this inquiry arises from the necessity for the LLMs to assess the length of the shortest path and discern as many qualifying paths as feasible. The objective of this task is to show the LLMs' capabilities in terms of fidelity and rectification.

\section{Prompt Engineering}
The prompt delivered to the LLMs plays a pivotal role in assessing model performance. Existing literature has introduced various prompting engineering methodologies, including techniques like chain-of-thought \cite{wei2022chain, kojima2022large}, few-shot prompting \cite{brown2020language}, least to most prompting \cite{zhou2022least}, and self-consistency \cite{wang2022self}. To comprehensively gauge the capabilities of LLMs in processing graph data, it is imperative to incorporate techniques from these established methodologies. As illustrated in Figure 1, our study employs a set of prompt engineering techniques. These techniques serve as a methodological tool to thoroughly investigate the capacities of LLMs in tackling problems related to graph structural data.

(1) \textbf{zero-shot prompting}: zero-shot prompting is arguably the most direct approach in the context of querying LLMs. For all problem sets, prompts under this category are manually constructed. A standard zero-shot prompt in this study consists of three distinct components: graph data, a query, and a tail.

Graph data can be represented through methods like adjacency matrices and adjacency lists. In this research, we employ a node list and an edge list to depict the graph data in plain text. By convention, both the node and edge lists are organized numerically. Given that edges are delineated in a sequential textual format, it is incumbent upon the LLMs to showcase proficiency in delineating the graph topology that these edges signify. Furthermore, we designate the graph as undirected. The edges inherently possess no directionality, implying that the terminal nodes are reciprocally connected.

Subsequent to the graph data, queries are clearly crafted. Typical queries posed to the LLMs might include explicit tasks like `find a single path from node \textit{p} to node \textit{q} connected by edges in the given graph' or `find a single wedge containing node 10 in the given graph'.

The end of the prompt includes a tail sentence. This stipulation, phrased as ``list the answers after `Ans:' in the format of [0-1-2]'', serves to standardize responses. Nevertheless, it is not mandatory to comply with this exact format in responses. Manual intervention remains necessary in extracting the conclusive results.

(2) \textbf{zero-shot chain-of-thought (CoT) prompting}: while chain-of-thought prompting \cite{wei2022chain} stimulates LLMs to show their reasoning process by presenting exemplars, zero-shot chain-of-thought prompting \cite{kojima2022large} is simply achieved by appending directives such as ``Let's think step by step'' in the end of the prompt. This technique is widely acknowledged to augment model performance. It prompts the LLMs to articulate their reasoning trajectory throughout the problem-solving phase, steering the model towards a methodical and systematic problem-solving strategy. Owing to its straightforwardness and efficacy, we have opted for zero-shot chain-of-thought prompting in this investigation. Unless specified otherwise, the terms ``chain-of-thought prompting'' and ``CoT'' in this study uniformly denote zero-shot chain-of-thought prompting.

(3) \textbf{few-shot prompting}: few-shot prompting\cite{logan2021cutting, lake2015human, brown2020language} essentially furnishes LLMs with exemplars to guide the generation of clearer responses. Different from the aforementioned CoT prompting, few-shot prompting tries to enhance model performance by presenting certain examples. This technique encourages LLMs to assimilate knowledge from the provided exemplars and formulate responses rooted in these exemplar outcomes. 

In this paper, we integrate a rudimentary exemplar of a similar problem with a standard zero-shot prompt. Such an approach enables the models to produce responses that are both contextually consistent and logically robust.

As a representative instance of few-shot prompting in this study, we present a foundational exemplar to the LLMs for addressing the problem of finding a simple path detailed as follows.

\textit{Q: Given the undirected graph with the specified nodes and edges, nodes: [0, 1, 2, 3, 4, 5, 6,], edges: [(0, 2), (0, 3), (0, 4), (0, 5), (0, 6), (1, 4), (1, 5), (2, 3), (2, 6), (3, 4), (3, 5), (3, 6), (4, 5), (5, 6),], find a single path from node 1 to node 6 connected by edges in the given graph, list the answer after ``Ans:'' in the format of [0-1-2].}

\textit{A :To find a single path from node 1 to node 6 in the given graph, we can use a depth-first search (DFS) algorithm. Here's the path: Start at node 1 $\to$ 5 $\to$ 0 $\to$ 2 $\to$ 6. Therefore, the path from node 1 to node 6 connected by edges in the given graph is: Ans: [1-5-0-2-6].}

\begin{figure*}[t]
\centering
\includegraphics[width=0.95\textwidth]{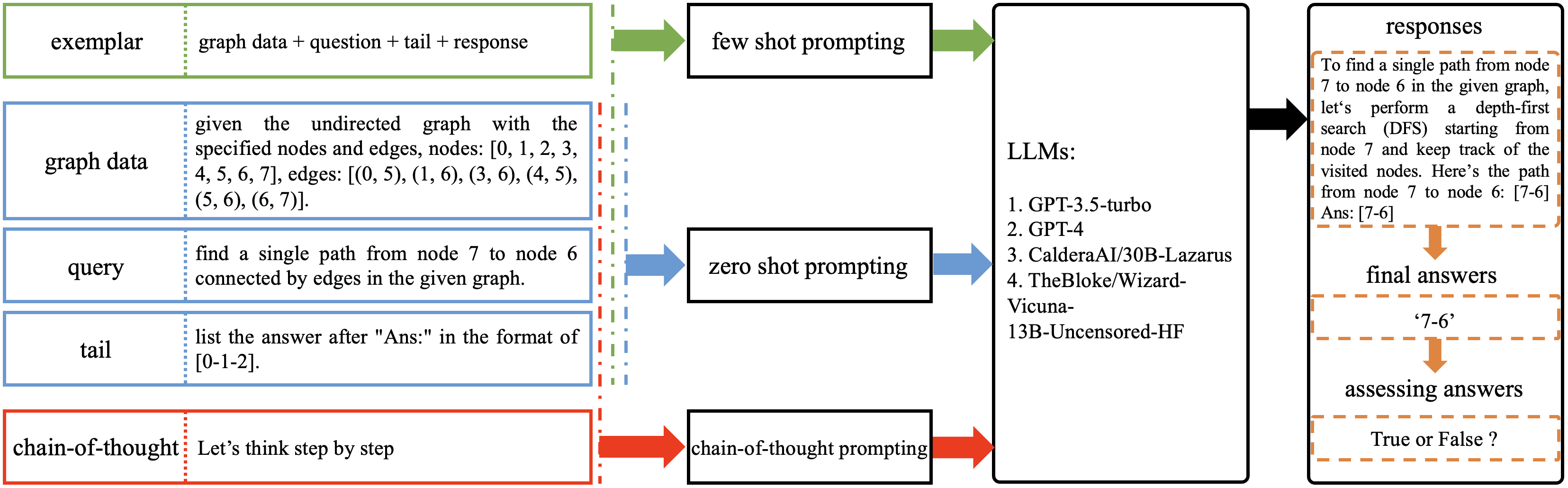}
\caption{Example of Different Prompts.}
\label{fig1}
\end{figure*}

\section{Experiments}

\subsection{Set-up}

The aforementioned graph problems are subjected to rigorous evaluation on a set of state-of-the-art LLMs, which include GPT-3.5-turbo \cite{ChatGPT}, GPT-4 \cite{ChatGPT, openaigpt4}, CalderaAI/30B-Lazarus \cite{30B-Lazarus} and TheBloke/Wizard-Vicuna-13B-Uncensored-HF \cite{Vicuna-13B, vicuna2023}. The selection of the latter two models is motivated by their high ranking on the Huggingface Open LLM Leaderboard \cite{huggingfacellm}. Unless otherwise specified, GPT-3.5-turbo and GPT-4 are collectively referred to as the ``GPT models'', while CalderaAI/30B-Lazarus and TheBloke/Wizard-Vicuna-13B-Uncensored-HF are designated as the ``open-source models''. Experiments are conducted on either a NVIDIA H100 80GB GPU or a NVIDIA A100 40GB GPU for the open-source models.

To circumvent the potential issue that the selected LLMs might already be trained with certain frequently used graphs, we employ a random graph generator to generate synthetic graphs. The evaluation dataset comprises 50 distinct graphs. Each graph may comprise between 20 to 50 nodes and 50 to 500 edges, with the edges being randomly generated.

In few-shot prompting, exemplars are derived from the prompts and the responses procured from the LLMs. Typically, these responses include both the reasoning logic and the eventual outcomes. Due to CalderaAI/30B-Lazarus and TheBloke/Wizard-Vicuna-13B-Uncensored-HF's suboptimal performance in zero-shot prompting scenarios, they utilize exemplars from GPT-3.5-turbo. In contrast, GPT-3.5-turbo and GPT-4 use self-generated exemplars.

\subsection{Metrics}

The objective of this study is to assess LLMs' understanding of graph data and their efficacy in addressing graph-related challenges. To thoroughly analyze our findings, we propose to dissect the results along four distinct dimensions: Comprehension, Correctness, Fidelity, and Rectification. 

The dimension of Comprehension pertains to the capacity of a model to decipher graph data represented in natural language. Correctness constitutes an evaluation metric for assessing the efficacy of language models in processing data and yielding accurate results. Fidelity illuminates the degree of confidence we can place in the results generated by these language models. Lastly, Rectification reflects the confidence LLMs exhibit in their responses, indicating their ability to correct errors while reinforcing accurate answers.

\section{Results}
Responses from the selected LLMs are collected and processed to extract clean answers. These answers are then evaluated based on four distinct evaluative metrics. The final results pertaining to the aforementioned queries can be found in Table 1.

\begin{table*}[t]
\centering
%%\resizebox{1.0\textwidth}{!}{
\begin{tabular}{ c  c  c  c  c}

  \multirow{2}{*}{Correct Ratio (\%)} & \multirow{2}{*}{GPT-3.5-turbo} & \multirow{2}{*}{GPT-4} & TheBloke/Wizard-Vicuna-13B & \multirow{2}{*}{CalderaAI/30B-Lazarus} \\
  & & & -Uncensored-HF & \\
 \hline
  \multicolumn{5}{c}{Connectivity (Find a Simple Path)} \\

 Zero-shot prompting & 50.0 & 72.0 & 4.0 &  4.0 \\ 

 CoT prompting & 50.0 & 68.0 & 2.0 & 6.0 \\ 

 Few-shot prompting & 54.0 & 76.0 & 6.0 & 0 \\ 
 \hline

  \multicolumn{5}{c}{Node Neighbor (Check the 1-hop Neighbor)} \\

 Zero-shot prompting & 60.4 & 91.2 & 0 &  9.6 \\ 

 CoT prompting & 62.8 & 92.0 & 24.8 & 34.4 \\

 Few-shot prompting & 60.0 & 88.0 & 25.2 & 32.0 \\ 
 \hline

  \multicolumn{5}{c}{Node Neighbor (Check the 2-hop Neighbor)} \\

 Zero-shot prompting & 80.8 & 88.4 & 0 &  8.0 \\ 

 CoT prompting & 73.2 & 91.6 & 32.8 & 28.0 \\

 Few-shot prompting & 66.0 & 91.2 & 14.4 & 8.0 \\ 
 \hline

  \multicolumn{5}{c}{Node Neighbor (Check the 3-hop Neighbor)} \\

 Zero-shot prompting & 81.6 & 86.8 & 0 &  11.2 \\ 

 CoT prompting & 75.2 & 90.0 & 27.6 & 45.2 \\

 Few-shot prompting & 75.2 & 78.0 & 9.6 & 29.2 \\ 
 \hline

  \multicolumn{5}{c}{Find Wedge (Centering the Target Node)} \\

 Zero-shot prompting & 76.0 & 94.0 & 0 &  0 \\ 

 CoT prompting & 78.0 & 88.0 & 4.0 & 0 \\

 Few-shot prompting & 76.0 & 86.0 & 12.0 & 0 \\
 \hline

  \multicolumn{5}{c}{Find Wedge (Containing the Target Node)} \\

 Zero-shot prompting & 92.0 & 94.0 & 0 &  0 \\ 

 CoT prompting & 96.0 & 98.0 & 0 & 12.0 \\

 Few-shot prompting & 96.0 & 94.0 & 28.0 & 8.0\footnotemark[1] \\
 \hline

  \multicolumn{5}{c}{Find Triangle} \\

 Zero-shot prompting & 60.0 & 76.0 & 0 &  18.0 \\ 

 CoT prompting & 60.0 & 74.0 & 6.0 & 6.0 \\

 Few-shot prompting & 72.0 & 84.0 & 16.0 & 6.0 \\
 \hline

  \multicolumn{5}{c}{Find Node Degree} \\
 
 Zero-shot prompting & 22.0 & 42.0 & 0 &  0 \\ 

 CoT prompting & 37.0 & 44.0 & 3.6 & 5.6 \\

 Few-shot prompting & 31.0 & 43.0 & 0 & 0.8 \\

\end{tabular}
%% }
\caption{Evaluations of LLMs on Graph Data} 
\label{table1}
\end{table*}

\subsection{Comprehension}

This metric evaluates the capabilities of LLMs in comprehending graph data. While graph data is presented in textual lists of nodes and edges, it differs from the sequential characteristic of natural language. It is incumbent upon LLMs to decipher and subsequently reconstruct the inherent topological structure from these plain texts. The aptitude to comprehend the interrelationships among the nodes in the graph forms the bedrock for problem-solving within the context of the input problems. Notably, this metric does not consider answer accuracy.

From this perspective, each of the four LLMs examined in this study exhibits a certain level of comprehension regarding the presented problems. Specifically, all the evaluated LLMs demonstrate a commendable degree of comprehension in multi-hop reasoning tasks. Using the connectivity (find a simple path) problem as a reference, Table 2 presents the comprehension ability of LLMs under zero-shot prompting. Specifically, if the LLMs produce a response that is relevant and logical with respect to the prompt, such a response is deemed to demonstrate comprehension and is classified as a positive response.

\begin{table}[ht]
\centering
%\resizebox{0.5\textwidth}{!}{
\begin{tabular}{c  c }

 model name & positive response ratio (\%) \\

 \hline
 GPT-3.5-turbo & 100 \\ 

 GPT-4 & 100 \\ 

 TheBloke/Wizard-Vicuna-13B & \multirow{2}{*}{78} \\

 -Uncensored-HF & \\

 CalderaAI/30B-Lazarus & 66 \\

\end{tabular}
% }
\caption{Ratio of positive responses indicative of comprehension in the connectivity problem}
\label{table2}
\end{table}

Despite the edges being supplied in a sequential text format, GPT models shows proficiency in structural reasoning over the provided edges. As a result, GPT models display enhanced understanding of graph topology when juxtaposed with the CalderaAI/30B-Lazarus and TheBloke/Wizard-Vicuna-13B-Uncensored-HF models. It's worth noting that these open-source models occasionally produce characters or words devoid of meaningful interpretation. In contrast to the occasional ambiguity manifested by the open-source models, GPT models consistently demonstrate coherence.

\footnotetext[1]{In the responses from this LLM, multiple answers may be produced. Of these responses, 26.0\% contain at least one accurate answer, while a mere 8.0\% are entirely correct.}

\subsection{Correctness}

A critical criterion, and perhaps the most consequential one, pertains to the correctness of the responses generated by the models. In our evaluations, both the CalderaAI/30B-Lazarus and TheBloke/Wizard-Vicuna-13B-Uncensored-HF models exhibit a strikingly low level of accuracy throughout these tests. A substantial improvement in correctness is observed in the case of both GPT models. It is rather intriguing to discover that despite the commendable performance of the two open-source models in disparate domains, their capacity to grasp the topological aspects of graphs is substantially constrained when compared against GPT models. These findings indicate that further endeavours are required to enhance the accuracy of the open-source models on graph-related problems.

In the two wedge-finding problems, we incrementally increase the topological reasoning requisites by specifying the structural position of the target nodes, leading to a noticeable decrease in model performance. Such findings underscore that graph structural reasoning remains a significant challenge for LLMs, including the sophisticated GPT models. As the topological reasoning demands of the queries intensify, LLMs exhibit a pronounced propensity for hallucination, leading to increased erroneous outputs.

The utilization of techniques such as zero-shot chain-of-thought and few-shot prompting does not unequivocally guarantee enhancements in the performance of the GPT models. In fact, such methods may potentially lead the models to generate fallacious responses, indicative of hallucination. Interestingly, GPT-4 appears to exhibit greater resilience to hallucination compared to GPT-3.5-turbo in certain tests. This observation suggests that enhanced reasoning capabilities in LLMs might aid in mitigating misleading effects during graph topological reasoning.

\subsection{Fidelity}

Considering the utilization of LLMs in graph contexts, it becomes critical to ascertain the extent of trust we can place in the responses generated by these models. Correctness constitutes an integral part of the overall trustworthiness of LLMs. Nevertheless, accurate answers often coexist with a multitude of erroneous ones in multi-answer scenarios, potentially inducing ambiguity. To capture the impact of these inaccurate outputs from LLMs, we have incorporated `fidelity' as a salient metric in our evaluation.

The GPT models outperform the other two models in our evaluations, thus rendering them much closer to realistic applications. Consequently, the following analysis primarily centers on the GPT models to assess the fidelity of their responses. A notable limitation is observed when GPT models tackle problems with multiple potential answers. They tend to generate a large quantity of spurious responses in contrast to accurate ones. This inclination tends to compromises the practical value of the answers produced by GPT models in resolving such types of graph-related problems. 

We propose a novel task wherein the GPT models are tasked with identifying all the shortest paths between a source node \textit{p} and a target node \textit{q} in the given graph. This task presents a heightened level of complexity compared to previous ones, necessitating the LLMs to both compute the shortest paths and enumerate every possible shortest route between the two specified nodes.

A typical zero-shot prompt for this task, along with its corresponding response, is illustrated subsequently.

\textit{Q: Given the undirected graph with the specified nodes and edges, nodes: [0, 1, 2, 3, ..., 35], edges: [(0, 2), (0, 5), (0, 20), ..., (32, 35)], find all the correct shortest paths from node 8 to node 22, list all the answers after ``Ans'' in the format of [0-1-2] and separate the answers by a comma.}

\textit{A: To find all the correct shortest paths from node 8 to node 22 in the given undirected graph, we can use a graph traversal algorithm like Breadth-First Search (BFS) or Dijkstra's algorithm. Here, I'll use BFS to find the shortest paths: Starting from node 8 (source), we explore the graph in a breadth-first manner, visiting neighbors before moving on to the next level. We maintain a queue to keep track of the nodes to be visited next, and we also maintain a list of visited nodes and their distances from the source. Here are the shortest paths from node 8 to node 22 in the given graph:}

\textit{Ans: [8-16-21-22], [8-16-25-21-22], [8-16-25-27-17-22], [8-16-25-27-17-20-22], [8-16-25-27-17-20-11-22], [8-16-25-27-17-4-20-22], [8-16-25-27-17-4-15-9-22], [8-16-25-27-17-4-15-9-17-22], [8-16-25-27-17-4-15-9-17-20-22], [8-16-25-27-17-4-15-9-17-20-11-22], [8-16-25-27-17-4-15-9-17-20-11-0-22], [8-16-25-27-17-4-15-9-17-20-11-0-5-22], [8-16-25-27-17-4-15-9-17-20-11-0-5-23-6-33-22], [8-16-25-27-17-4-15-9-17-20-11-0-5-23-6-2-22]}

\textit{Please note that there may be multiple correct shortest paths in a graph, and the above paths are the shortest paths found using BFS based on the provided graph structure.}

The results derived from our evaluation dataset are presented in Table 3. It is evident that both GPT models occasionally produce erroneous responses. Remarkably, GPT-3.5-turbo produces erroneous responses at a rate 7.57 times higher than that of correct answers. This phenomenon reduces the fidelity of GPT models when addressing more complex graph-related tasks.

\begin{table}[ht]
\centering
%% \resizebox{0.5\textwidth}{!}{
\begin{tabular}{c  c  c  c  c}

 all shortest paths & correct answers & erroneous answers & correct ratio & comparison ratio\footnotemark[2] \\

 \hline
 GPT-3.5-turbo & 107 & 810 & 24.65\% &  7.57 \\ 

 GPT-4 & 124 & 121 & 28.57\% & 0.98 \\ 

\end{tabular}
%% }
\caption{Comparison of Correct and Erroneous Answers Generated by GPT models}
\label{table3}
\end{table}

\subsection{Rectification}

In terms of the fourth metric, rectification reflects the level of confidence exhibited by the GPT models in their generated responses, covering both correct and incorrect outputs. Similar to real-world decision-makers who might make both accurate and erroneous judgments, an LLM with optimal rectification should emphasize its accurate outputs while rectifying its mistakes, mirroring the behavior of a well-calibrated individual.

\footnotetext[2]{Comparison ratio is defined as the quotient obtained from dividing the number of erroneous answers by the number of correct answers.}

We expand upon the task of identifying all the shortest paths between a source node \textit{p} and target node \textit{q}, as delineated in the preceding section. Upon the GPT models' initial response generation, a secondary query is introduced, tasking the LLMs with the removal of inaccuracies from the potential shortest paths identified initially. Using the responses from GPT-3.5-turbo's and GPT-4's initial query, both GPT-3.5-turbo and GPT-4 are then prompted with the secondary task.

\begin{table}[ht]
%% \resizebox{0.5\textwidth}{!}{
\centering
\begin{tabular}{c  c  c  c }

 initial model & GPT-3.5-turbo & GPT-3.5-turbo &  GPT-4 \\

 verifying model & GPT-3.5-turbo &  GPT-4 & GPT-4  \\

 \hline
 correct answers & 107 & 83 & 93  \\ 

 erroneous answers & 806 & 144 & 48  \\ 
 
 correct ratio &  24.65\% & 19.12\% & 21.43\% \\ 
 
 comparison ratio\footnotemark[2] & 7.53 & 1.73 & 0.52 \\ 
 
 new answers & 0 & 4 & 1 \\ 

\end{tabular}
%% }
\caption{Comparison of Correct and Erroneous Answers Generated by GPT models after Re-evaluations}
\label{table4}
\end{table}

In Table 4, it is evident that GPT-3.5-turbo rectify merely four erroneous answers it has initially produced. Conversely, GPT-4 manages to rectify over half of its prior erroneous answers and addresses 82\% of the inaccuracies originating from GPT-3.5-turbo. Nonetheless, it is noteworthy that GPT-4 concurrently eliminates 26\% of the correct outputs previously produced by both GPT-3.5-turbo and its own earlier iterations.

The observations indicate that once GPT-3.5-turbo has formulated its reasoning logic, it displays high confidence in such logic, even in instances where the responses are incorrect. It presents a challenge to prompt GPT-3.5-turbo to self-filter incorrect responses. It is possible that GPT-4 possesses the capability to exclude certain erroneous answers generated both by GPT-3.5-turbo and its own earlier iterations. Such a finding might imply that a GPT model with superior reasoning capability might exhibit a certain degree of rectification proficiency over responses produced by both its predecessor and its own previous iterations. Nonetheless, the propensity of GPT-4 to discard a substantial proportion of accurate responses leads to a marked decrease in the correctness ratio upon subsequent evaluation.

Interestingly, our results indicate that GPT-4 demonstrates a modest inclination to generate additional responses beyond those initially produced by GPT-3.5-turbo and its own prior outputs. Notably, the five answers that emerged are all accurate shortest paths relevant to their respective queries. It should be highlighted that certain prompts might exceed the token limitation established for GPT-4, leading to the hypothesis that GPT-4 refrains from adjusting these responses in subsequent query rounds.

\section{Comparative Study}
We subsequently conduct a comparative analysis to show the performance differentials between the two GPT models in relation to graph size, graph sparsity, and the variance in examples employed for few-shot prompting.

\subsection{Positive and negative few-shot exemplars}
It is noteworthy that in certain problems, exemplars can be presented in either a positive or negative context. However, the implications of employing diverse exemplars have not been extensively explored using graph data. To show the impact of varying exemplars on few-shot prompting, we expand the experimental settings for the node neighbor classification problem.

Regarding the node neighbor classification problem, exemplars can be presented in two distinct orientations. In a positive framework, exemplars can be offered by prompting the GPT models to determine if a node serves as the \textit{k}-hop neighbor of a designated node, with the ensuing response being ``Yes''. The exemplar can be structured as follows:

\textit{Q: Given the undirected graph with the specified nodes and edges, nodes: [0, 1, 2, 3, 4, 5, 6, 7], edges: [(0, 5), (1, 6), (3, 6), (4, 5), (5, 6), (6, 7)]. Is node 5 the 1-hop neighbor of node 4? List the answers after ``Ans:'' in the format of [Yes, No,]. }

\textit{A: To determine if node 5 is a 1-hop neighbor of node 4 in the given undirected graph, we need to check if there is a direct edge between node 4 and node 5. Looking at the list of edges provided, we can see that there is an edge between node 4 and node 5, specifically (4, 5). Therefore, node 5 is indeed a 1-hop neighbor of node 4. Ans: [Yes,].}

In the negative context, we format an exemplar by prompting the GPT models to identify a prospective neighbor node of a target node, where the desired answer is ``No''. The exemplar can be delineated as follows:

\textit{Q: Given the undirected graph ... Is node 6 the 1-hop neighbor of node 4? List the answers after ``Ans:'' in the format of [Yes, No,]. }

\textit{A: To determine if node 6 is a 1-hop neighbor of node 4, we need to check if there exists an edge between them directly. Let's examine the given edges: Edges: [(0, 5), (1, 6), (3, 6), (4, 5), (5, 6), (6, 7)]. The edge (4, 5) exists in the given list, but there is no direct edge between node 4 and node 6. Therefore, node 6 is not a 1-hop neighbor of node 4. Ans: [No,].}

As delineated in Table 1, all the exemplars employed in the few-shot prompting are positive exemplars, signifying that the queried nodes are indeed the \textit{k}-hop neighbors of the target nodes. Subsequently, we evaluate the GPT models using negative exemplars to assess their performance. The outcomes are presented in Table 5.

\begin{table}[t]
\centering
% \resizebox{0.5\textwidth}{!}{
\begin{tabular}{c c c }
 Correct Ratio(\%) & GPT-3.5-turbo & GPT-4 \\

  \hline
 check the 1-hop neighbor  & 62.0 & 84.4 \\ 

 check the 2-hop neighbor  & 85.2 & 73.2 \\ 

 check the 3-hop neighbor & 62.8 & 81.6 \\ 
\end{tabular}
% }
\caption{Few-shot prompting on node neighbor classification problems using negative exemplars}
\label{table5}
\end{table}

\begin{table}[t]
\centering
% \resizebox{0.5\textwidth}{!}{
\begin{tabular}{c c c }
 Answer distributions (\%) & True & False \\
 \hline
 Node Neighbor (Check the 1-hop Neighbor) & 58 & 42 \\ 
 Node Neighbor (Check the 2-hop Neighbor) & 87.2 & 12.8 \\ 
 Node Neighbor (Check the 3-hop Neighbor) & 38 & 62 \\ 
\end{tabular}
% }
\caption{Distributions of True and False answers in node neighbor classification problems}
\label{table6}
\end{table}

It is evident that distinct exemplars significantly influence the performance of GPT models. In the task of classifying 2-hop neighbor nodes, GPT-3.5-turbo demonstrates enhanced performance when presented with a negative exemplar compared to a positive one; conversely, GPT-4 exhibits the inverse trend.

\subsection{Graph Size}

In this section, our primary objective is to investigate the impact of graph sizes on the performance of the two GPT models. To mitigate the potential effect of the number of edges, we stochastically generate the edges within the graph, ensuring that their quantity approximates half of the maximum potential edge count. We persist in employing the queries related to the all shortest paths problem to prompt the GPT models. The outcomes are presented in Table 7.

\begin{table}[ht]
\centering
%% \resizebox{0.5\textwidth}{!}{
\begin{tabular}{c  c  c  c  c}

 model name & correct answers & erroneous answers & correct ratio & comparison ratio\footnotemark[2] \\

 \hline
  \multicolumn{5}{c}{Node Number : 10} \\
 GPT-3.5-turbo & 18 & 6 & 46.15\% &  0.33 \\ 
 GPT-4 & 32 & 0 & 82.05\% & 0 \\ 

 \hline
  \multicolumn{5}{c}{Node Number : 15} \\
 GPT-3.5-turbo & 13 & 7 & 32.50\% &  0.54 \\ 
 GPT-4 & 36 & 2 & 90.00\% & 0.06 \\ 

 \hline
  \multicolumn{5}{c}{Node Number : 20} \\
 GPT-3.5-turbo & 13 & 282 & 29.55\% &  21.69 \\ 
 GPT-4 & 25 & 3 & 56.82\% & 0.12 \\ 

 \hline
  \multicolumn{5}{c}{Node Number : 25} \\
 GPT-3.5-turbo & 33 & 335 & 55.00\% &  10.15 \\ 
 GPT-4 & 24 & 0 & 40.00\% & 0 \\ 

 \hline
  \multicolumn{5}{c}{Node Number : 30} \\
 GPT-3.5-turbo & 44 & 359 & 55.70\% &  8.16 \\ 
 GPT-4 & 27 & 28 & 34.18\% & 1.04 \\ 

  \hline
  \multicolumn{5}{c}{Node Number : 35} \\
 GPT-3.5-turbo & 35 & 2668 & 46.05\% &  76.23 \\ 
 GPT-4 & 18 & 197 & 23.68\% & 10.94 \\ 
 
\end{tabular}
%% }
\caption{All shortest paths problem on graphs with different sizes}
\label{table7}
\end{table}

It is noteworthy to observe that as the number of nodes increases, the GPT models exhibit an escalating tendency to produce erroneous answers. This observation intimates that larger graphs might engender a more pronounced susceptibility to hallucination compared to their smaller counterparts. Furthermore, GPT-3.5-turbo appears to maintain consistently adequate performance until the node count approaches 20, whereas GPT-4 demonstrates commendable efficacy up to a threshold of 35 nodes.

\subsection{Graph Sparsity}
Subsequently, our analysis aims to show the influence of graph sparsity on the performance of the GPT models. As evidenced by the data presented in Table 7, all graphs generated for this section maintain a uniform node count of 25, with edge numbers approximating a specific proportion of the maximum possible edge count. The investigation persists utilizing the queries from the all shortest paths problem to prompt the GPT models. The outcomes of this evaluation can be found in Table 8.

\begin{table}[t]
\centering
%% \resizebox{0.5\textwidth}{!}{
\begin{tabular}{c  c  c  c  c}

 model name & correct answers & erroneous answers & correct ratio & comparison ratio\footnotemark[2] \\

 \hline
  \multicolumn{5}{c}{Edge Percentage : 10\%} \\
 GPT-3.5-turbo & 2 & 20 & 5.41\% &  10.00 \\ 
 GPT-4 & 26 & 0 & 70.27\% & 0 \\ 

 \hline
  \multicolumn{5}{c}{Edge Percentage : 30\%} \\
 GPT-3.5-turbo & 4 & 40 & 8.33\% &  10.00 \\ 
 GPT-4 & 19 & 8 & 39.58\% & 0.42 \\ 

 \hline
  \multicolumn{5}{c}{Edge Percentage : 50\%} \\
 GPT-3.5-turbo & 37 & 1828 & 54.41\% &  49.41 \\ 
 GPT-4 & 28 & 4 & 41.18\% & 0.14 \\ 

 \hline
  \multicolumn{5}{c}{Edge Percentage : 70\%} \\
 GPT-3.5-turbo & 81 & 2088 & 69.23\% &  25.78 \\ 
 GPT-4 & 39 & 14 & 33.33\% & 0.36 \\ 

 \hline
  \multicolumn{5}{c}{Edge Percentage : 90\%} \\
 GPT-3.5-turbo & 104 & 583 & 55.32\% &  5.61 \\ 
 GPT-4 & 60 & 17 & 31.91\% & 0.28 \\ 

\end{tabular}
%% }
\caption{All shortest paths problem on graphs with different sparsity}
\label{table8}
\end{table}

The outcomes suggest that both GPT models exhibit enhanced performance in identifying accurate answers on denser graphs. Yet, as the number of edges escalates, the GPT models exhibit an increasing propensity to produce erroneous outputs. Notably, GPT-3.5-turbo might identify a greater number of accurate answers compared to GPT-4; however, it concomitantly produces a significantly higher number of spurious responses. This tendency towards hallucination becomes pronounced once the graph density surpasses a specific threshold.

\section{Related Work}

\subsection{LLMs for solving graph based problems}
Dissimilar to the way that other machine learning models dealing with graph data, LLMs process natural language inputs and generate plain text responses. Though LLMs have undergone extensive evaluations in various fields using multiple metrics \cite{li2023evaluating}, assessing their performance with graph data remains a critical avenue of inquiry. Prior investigations by Guo et al. \cite{guo2023gpt4graph} tested a variety of common graph tasks utilizing LLMs, inclusive of text-davinci-003. Their findings revealed that the performance of the selected LLMs fell short of achieving a thorough understanding of graph data. Wang et al. \cite{wang2023can}, on the other hand, discerned that language models do exhibit initial abilities in graph reasoning and proceeded to construct a benchmark comprised of several graph-related tasks. Zhang et al. \cite{zhang2023graph} introduced a novel framework, termed ``Graph-ToolFormer,'' designed to enable LLMs to self-instruct using prompts enhanced by ChatGPT, facilitating the utilization of external graph reasoning API tools. The focus of the our study pivots towards uncovering the capacity of LLMs to resolve specified graph-related issues, with particular emphasis on structural reasoning abilities.

\subsection{LLMs and knowledge graphs}
Knowledge graphs are structured representations that encapsulate rich factual knowledge \cite{ji2021survey}. Conversely, LLMs have faced critique for their perceived inability to precisely encapsulate factual knowledge due to their black-box nature \cite{wang2023robustness, ji2023survey}. A substantial body of research \cite{yin2022survey} has delved into the potential synergy between knowledge graphs and LLMs. Some earlier studies suggested the integration of knowledge graphs into both the pre-training and inference stages of LLMs to imbue them with factual knowledge \cite{zhang2019ernie, liu2021kg}. Conversely, other studies have highlighted the utility of LLMs in knowledge graph-associated tasks, including semantic embedding and text generation \cite{wang2021kepler, ke2021jointgt}. Additional research underscored the collaborative potential between LLMs and knowledge graphs, which can bolster performance in areas such as knowledge graph reasoning \cite{wang2023unifying}. Furthermore, Pan et al. \cite{pan2023unifying} have proposed a comprehensive roadmap, detailing several frameworks aimed at harmonizing LLMs with knowledge graphs. The integration of knowledge graphs underscores potential future applications of LLMs with certain graph datasets. Nevertheless, in our present study, we maintain a focus on the assessment of LLMs utilizing conventional graph data.

\section{Conclusions}

In the present study, we undertake a comprehensive investigation aimed at evaluating the capacity of several LLMs to process graph data. The findings are indeed striking, revealing that GPT models are capable of delivering high-quality responses to problems pertaining to graph data. These models exhibit a substantially superior ability to comprehend and process topological data associated with graphs in comparison to the other two counterparts. Such observations illuminate the potential of LLMs in managing structural data. However, substantial effort is warranted to ensure satisfactory correctness. A discernible limitation lies in the unwarranted confidence exhibited by GPT models concerning their responses, even when such responses may be erroneous. This necessitates increased importance in identifying and rectifying such answers. Furthermore, when engaging with structural data, common enhancement techniques such as zero-shot chain-of-thought or few-shot prompting may not yield the anticipated benefits. These methods may precipitate substantial hallucinations, leading to the generation of incorrect answers. The phenomena become increasingly discernible when the demands for understanding graph topology are amplified. We further investigate the competencies of the two GPT models using fidelity and rectification as evaluative metrics. Regrettably, neither of the GPT models show optimal performance in our assessments.

%Bibliography
\bibliographystyle{unsrt}  
\bibliography{references}  

\end{document}